%% file: Article_V_-_I.tex
\documentclass[10pt, reqno]{amsart}

\usepackage[utf8]{inputenc}
\usepackage{amsmath}
\usepackage{amsthm}
\usepackage{accents}
\usepackage{enumerate}
\usepackage{amsfonts}
\usepackage{verbatim}
\usepackage[foot]{amsaddr}
\usepackage{graphicx}
\usepackage{float}
\usepackage[ font=footnotesize ]{subfig}
\usepackage[margin=3.5cm]{geometry}
\usepackage{tikz}
\usetikzlibrary{matrix,chains,positioning,decorations.pathreplacing,arrows}

\pdfoutput=1 
\theoremstyle{plain}
\newtheorem{theorem}{Theorem}

\theoremstyle{remark}

\theoremstyle{definition}

\begin{document}
	
	       \author[Rosengren]{Sebastian Rosengren }      
	       \address{Department of Mathematics, Stockholm University, 106 91 Stockholm, Sweden.}
	       \email{rosengren@math.su.se}

\title[Shape Theorems --- Predictability and New Applications]{Predicting First Passage Percolation Shapes Using Neural Networks} 
\maketitle

\begin{abstract}
	
	Many random growth models have the property that the set of discovered sites, scaled properly, converges to some deterministic set as time grows. Such results are known as shape theorems.
	Typically, not much is known about the shapes. For first passage percolation on $\mathbb{Z}^d$ we only know that the shape is convex, compact, and inherits all the symmetries of $\mathbb{Z}^d$.
	Using simulated data we construct and fit a neural network able to adequately predict the shape of the set of discovered sites from the mean, standard deviation, and percentiles of the distribution of the passage times.
	The purpose of the note is two-fold. The main purpose is to give researchers a new tool for \textit{quickly} getting an impression of the shape from the distribution of the passage times --- instead of having to wait some time for the simulations to run, as is the only available way today.
	The second purpose of the note is simply to introduce modern machine learning methods into this area of discrete probability, and a hope that it stimulates further research.

\end{abstract}
\newpage

\section{Introduction}
First passage percolation is a well-studied random growth model on $\mathbb{Z}^d$. It was first introduced in \cite{Hammersley1965}
as a model for how a liquid flows through a random medium, and a recent summary is provided in \cite{auffinger201550}.

\begin{figure}[H]
	\centering
	\includegraphics[scale=0.25]{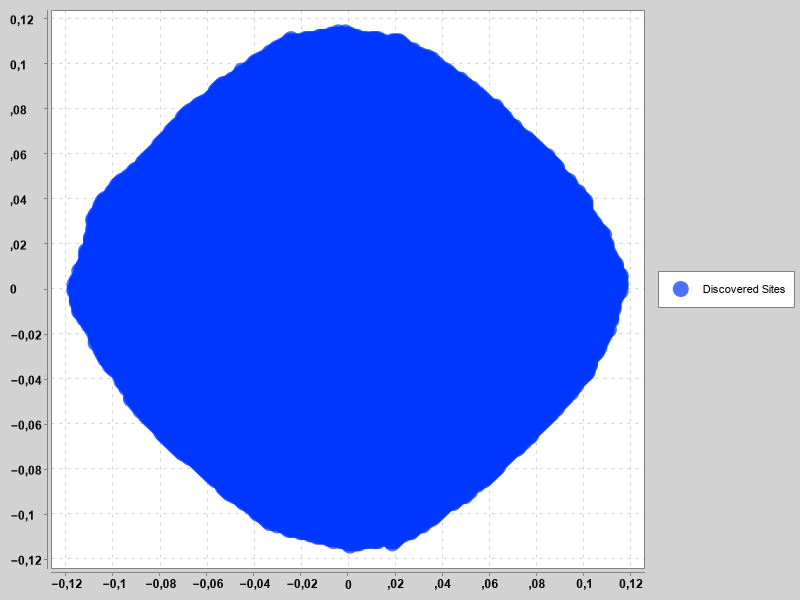}
	\caption{Simulation of $B(t) / t$ with  $\tau_e \sim \Gamma(n = 10, \lambda = 1)$ for $t = 1000.$}
	\label{fig:sub1}
\end{figure}

The model is defined as follows (we will follow \cite{auffinger201550} in notation). On each edge $e$ in $\mathbb{Z}^d$ we place a non-negative random variable $\tau_e$ called the \textit{passage time} of the edge. The family of random variables $\{\tau_e \}$ is assumed to be i.i.d. In the original setting of the model, $\tau_e$ is interpreted as the time it takes for the liquid to pass through the edge $e$. A key concept is that of a path $\Gamma$, defined as a sequence of edges $e_1, e_2, \ldots$ such that $e_n$ and $e_{n+1}$ are connected. We define the \textit{passage time} of a path $\Gamma$ as
\begin{align*}
	T(\Gamma) = \sum_{e \in \Gamma } \tau_e,
\end{align*}
i.e. the time it takes to traverse the path.
Furthermore, we define the passage time between two \textit{points} $x,y \in \mathbb{Z}^d$ to be
\begin{align*}
	T(x,y) = \inf\{ T(r):\ r \text{ is a path from }x\text{ to }y \}.
\end{align*}
Let
\begin{align*}
	B(t) = \{ y\in \mathbb{Z}^d:\ T(0,y)\leq t \}
\end{align*}
i.e. the set of vertices that can be reached from the origin by time $t$ --- here referred to as the set of \textit{infected sites}. It turns out that for a wide range of passage times the set of infected sites, properly scaled by time, converges to a deterministic set. This type of results are called \textit{shape theorems}. Let $\bar{B}(t) = \{ x+[-\frac{1}{2},\frac{1}{2} ]^d:\ x\in B(t) \}$ be the continuum version of $B(t)$.
\begin{theorem}[Cox and Durrett \cite{cox1981}]
	\label{shape}
	Assume that $\{\tau_e \}$ satisfies
	\begin{enumerate}[(i)]
		\item $\mathbb{E}(\min\{ \tau_{e_1}, \tau_{e_2},\ldots, \tau_{e_{2d}}\} ) < \infty$, where $\tau_{e_1},\tau_{e_2},\ldots, \tau_{e_{2d}}$ are iid copies of $\tau_e$.
		\item $\mathbb{P}(\tau_e = 0) < p_c(d)$ where $p_c(d)$ is the threshold for bond percolation on $\mathbb{Z}^d$.
	\end{enumerate}
Then, there exists a convex non-empty compact set $\mathcal{B} \in \mathbb{R}^d$ such that for each $\epsilon>0$, 
\begin{align*}
	\mathbb{P}\left(   (1-\epsilon)\mathcal{B} \subset \frac{\bar{B}(t)}{t} \subset (1+\epsilon)\mathcal{B} \text{ for all large }t \right)  = 1.
\end{align*}

\end{theorem}
Figure \ref{fig:sub1} gives an impression of $\mathcal{B}$ for gamma distributed passage times.
Not much is known about the set $\mathcal{B}$ other than that it is non-empty, convex, compact, and inherits all symmetries of $\mathbb{Z}^d$. Determining the shape is a well-studied but difficult problem, see e.g. \cite{DeijfenAlm} and the references therein. Currently, for non-degenerate distributions, the only way to go from the distribution of $\tau_e$ to $\mathcal{B}$ is through simulation. 

There is nothing wrong with this, but in many situations waiting for simulations can be quite disruptive of the work flow --- as they take time to run, making it difficult to test out ideas in a timely fashion. Furthermore, simulations gives little insight in the functional relationship between the distribution of $\tau_e$ and the shape $\mathcal{B}$.
 
The purpose of this note is mainly to mitigate the first problem in $d=2$. This is done by simulating first passage percolation on $\mathbb{Z}^2$, for a variety of passage times (belonging to the normal, gamma, or beta distribution), and then fitting a neural network able to \textit{approximate} the shape of $\mathcal{B}$. The model uses the percentiles (excluding the 0th and 100th percentile), the mean, the standard deviation, and an $x$-coordinate as input and predicts a $y$-coordinate (see Section \ref{useage} for further details). Neural networks are suitable for the problem, mainly because they are flexible enough to approximate a large class of functions, see Section \ref{nn}. More specifically, neural networks can, given enough data, capture non-linear dependencies of explanatory variables ($\mu_e, \sigma_e, x, q_{0.01}, \ldots, q_{0.99}$) on the response ($y$) \textit{indirectly}, i.e. we do not have to specify the dependencies. This is suitable here since so little is known about the shapes.
We shall see that the method works rather well, and that the generalizing capabilities are promising. 
 
\subsection{Raison d'être --- Intended Use}
The intended user would be a researcher that \textit{quickly} wants a sense of the shape $\mathcal{B}$, from some easily accessible properties of the underlying distribution. The operating word here is quickly, as simulations can already provide an accurate picture of the final shape. However, simulations tend to take time to run and in many cases, it might be better to have something (slightly) less accurate but faster.

\subsection{How to use the models}
\label{useage}
Again, the idea of the note is to be able to take some information from the distribution of $\tau_e$ and from this predict the shape $\mathcal{B}$. Hence, we have to decide how to summarize the information in the distribution of $\tau_e$. There are of course many ways to do this, but we have settled on the percentiles $1\%-99\%$ $(q_{0.01}-q_{0.99})$, the mean $(\mu_e)$, and the standard deviation ($\sigma_e$). These quantities are easy to calculate, and more importantly they seem to work well for modeling. As mentioned before, the passage times are allowed to belong to one of three distributions: normal (conditioned on being positive), gamma, or (scaled) beta. The reasons for choosing these distributions is convenience --- they are well-known distributions that can be simulated efficiently, and the distributions  are fairly different. The normal and gamma distribution have unbounded support, while the beta distribution has bounded support. All distributions however have support down to 0, so we exclude the 0th percentile since it always takes value 0 (so it contains no information). We also the 100th percentile which can be infinitely large. Hence, we are only studying shapes that are generated by absolutely continuous distributions, and therefore model predictions should be restricted to shapes generated by passage times from this class of distribution.

Furthermore, we require an $x$-value to predict the corresponding $y$-value. Since we know that the shape inherits all symmetries from $\mathbb{Z}^2$ it is enough if the model can predict the shape for points $(x,y)$ in the first quadrant that lies above the line $y = x$ (see Section \ref{transformation} for further details). Hence, our model is of the form
\begin{align*}
y = f(x, \mu_e, \sigma_e, q_{0.01},q_{0.02},\ldots, q_{0.99}).
\end{align*}
Note that for a given distribution, $\mu_e, \sigma_e, q_{0.01},q_{0.02},\ldots, q_{0.99}$ are fixed, and only $x$ is allowed to vary.

It may seem counter-intuitive that the model takes a $x$-coordinate as input since it is part of the shape $ \mathcal{B}$, which is to be predicted. For instance, for a given distribution we do not know a priori which $x$-values are "valid", i.e. which $x\in \mathcal{B}$.
But we can solve this by predicting in an iterative fashion. We always know that $x=0$ is part of the function domain. After this we simply try new slightly larger $x$-values until we get a $y$-value that lies below the line $y = x$. 

\subsection{Results}
After comparing model performances on training and test data, we pick as a final model a neural net with ten layers, 60 hidden units, and ReLu activation (see Section \ref{nn} for definitions). Using the mean absolute percentage error as a metric ($ \text{MAPE} =  \sum_{i=1}^{n}\frac{|y_i-\hat{y}_i|}{y_i}  / n$) we have:
\begin{itemize}
	\item a 1.72 \% mean absolute percentage error on training data ($\mu_e \in [1,20]$).
	\item a 3.09 \% mean absolute percentage error on test data ($\mu_e \in [25,45]$).
\end{itemize}

\begin{figure}[H]
	\centering
	\subfloat[Normal Distribution $\mu_e = 10.03$]{\includegraphics[width=.375\textwidth]{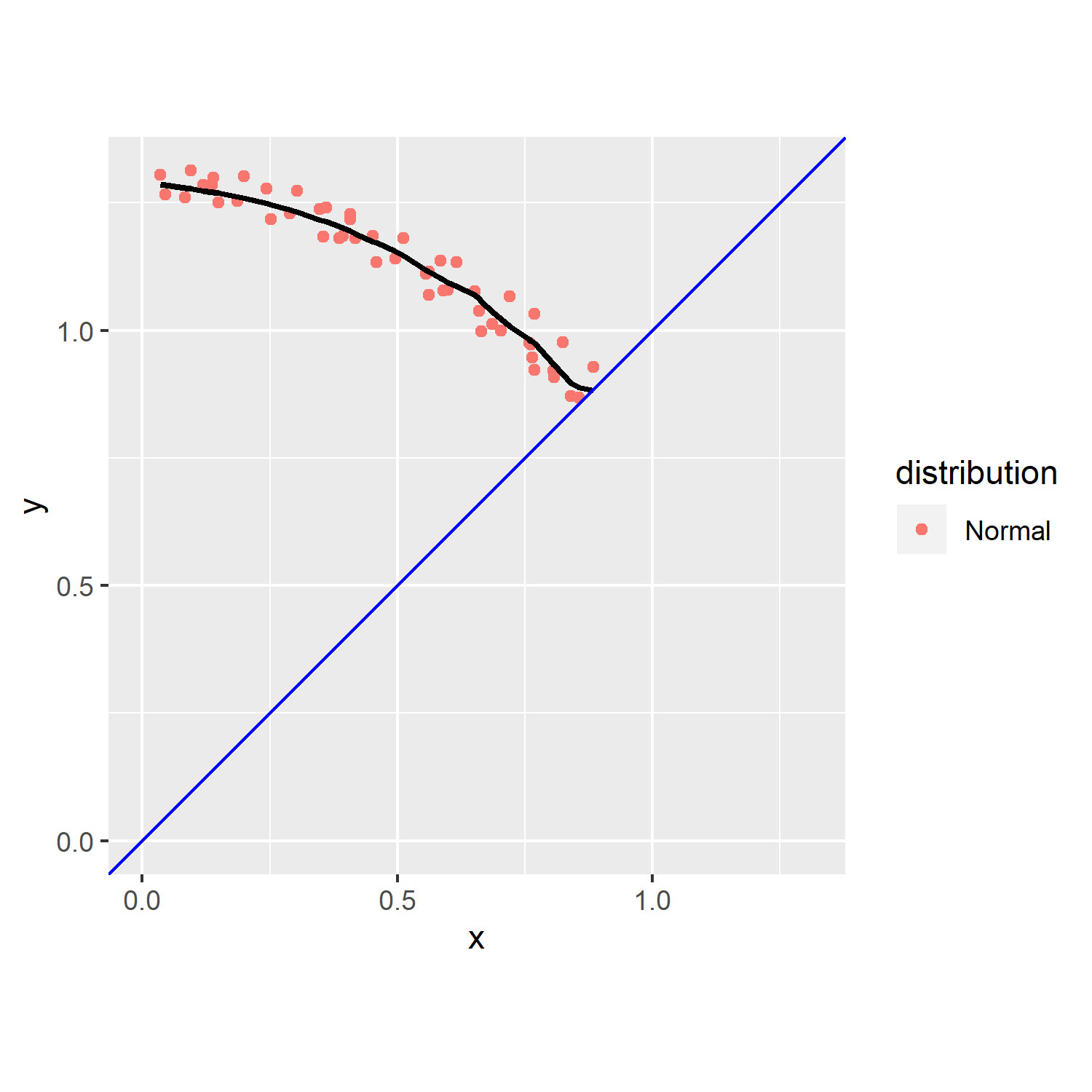}}\quad
	\subfloat[Gamma Distribution $\mu_e = 8.32$]{\includegraphics[width=.375\textwidth]{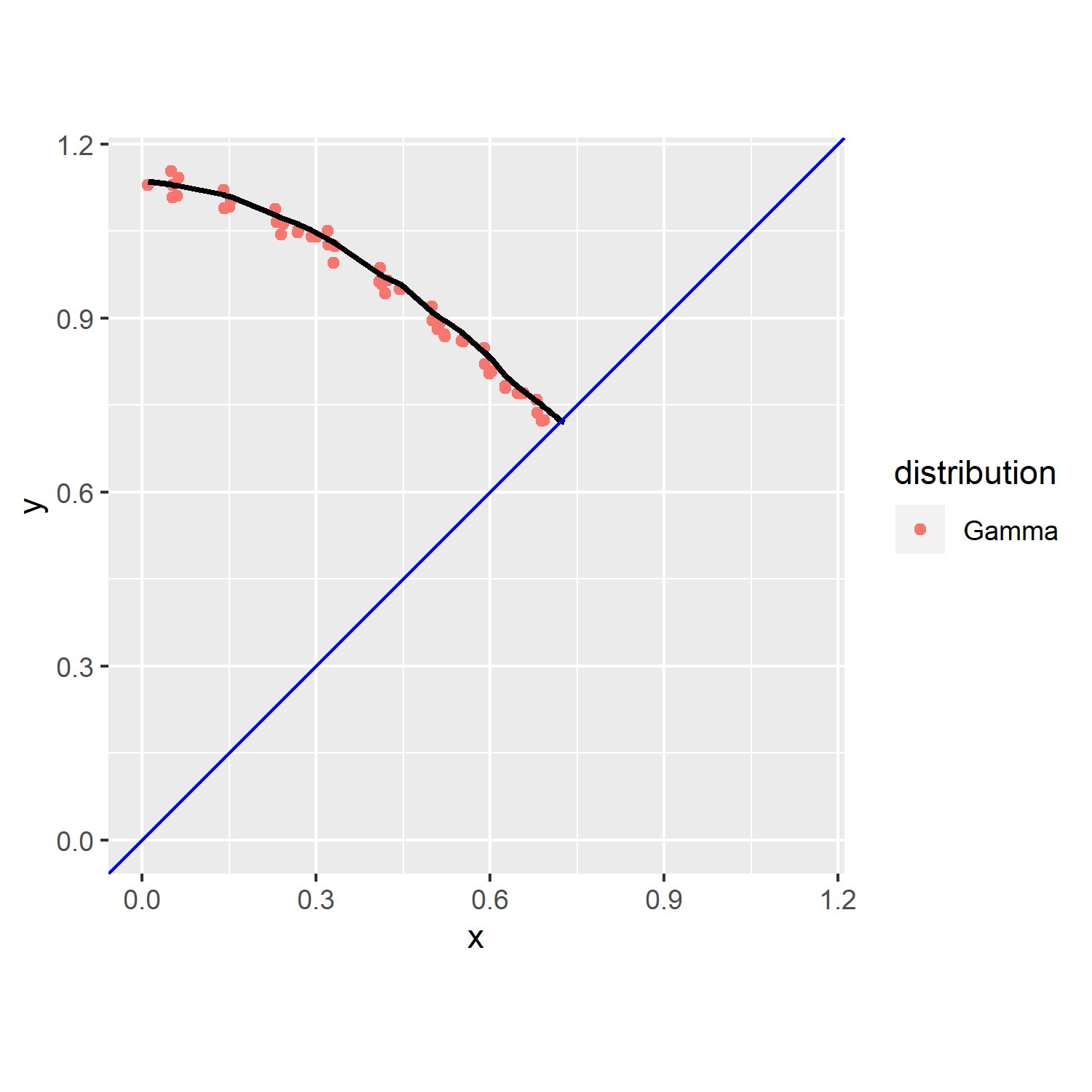}}\\
	\subfloat[Beta Distribution $\mu_e = 8.39$]{\includegraphics[width=.375\textwidth]{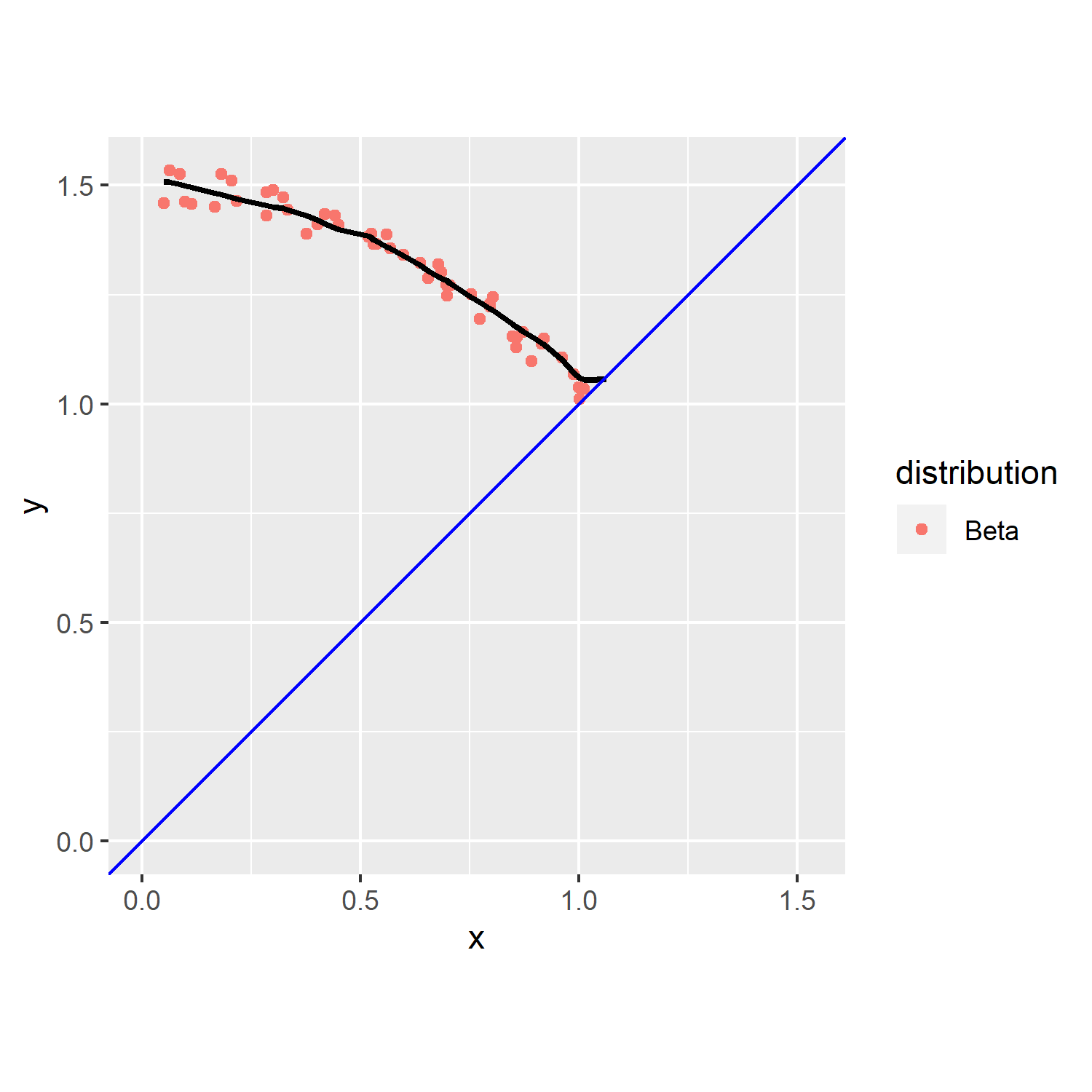}}\quad
	\caption{Illustration of neural network model predicting shape for \textit{training data}, for three representative distributions}
	\label{train}
\end{figure}

\begin{figure}[H]
	\centering
	\subfloat[Normal Distribution $\mu_e = 35.04$]{\includegraphics[width=.375\textwidth]{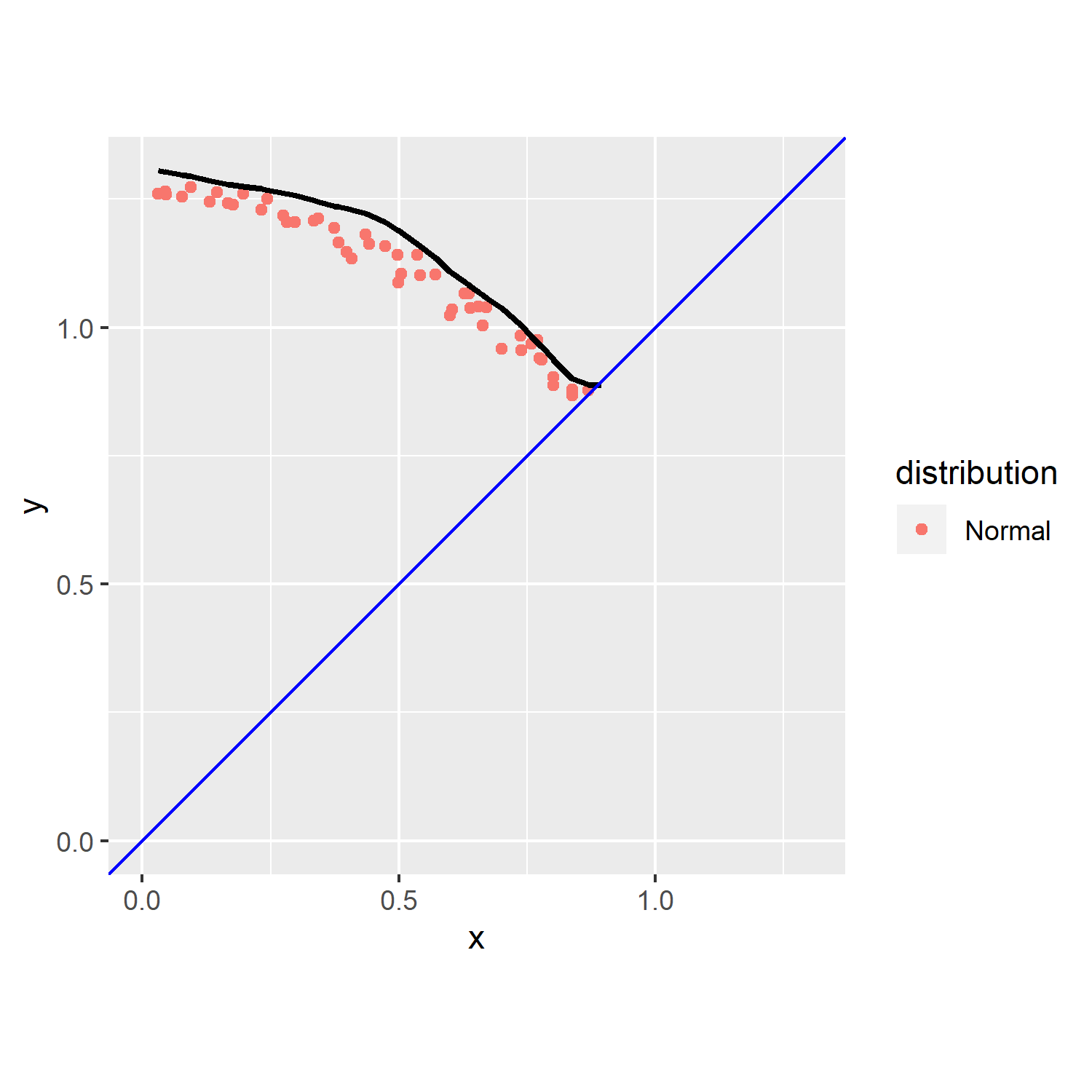}}\quad
	\subfloat[Gamma Distribution $\mu_e = 34.97$]{\includegraphics[width=.375\textwidth]{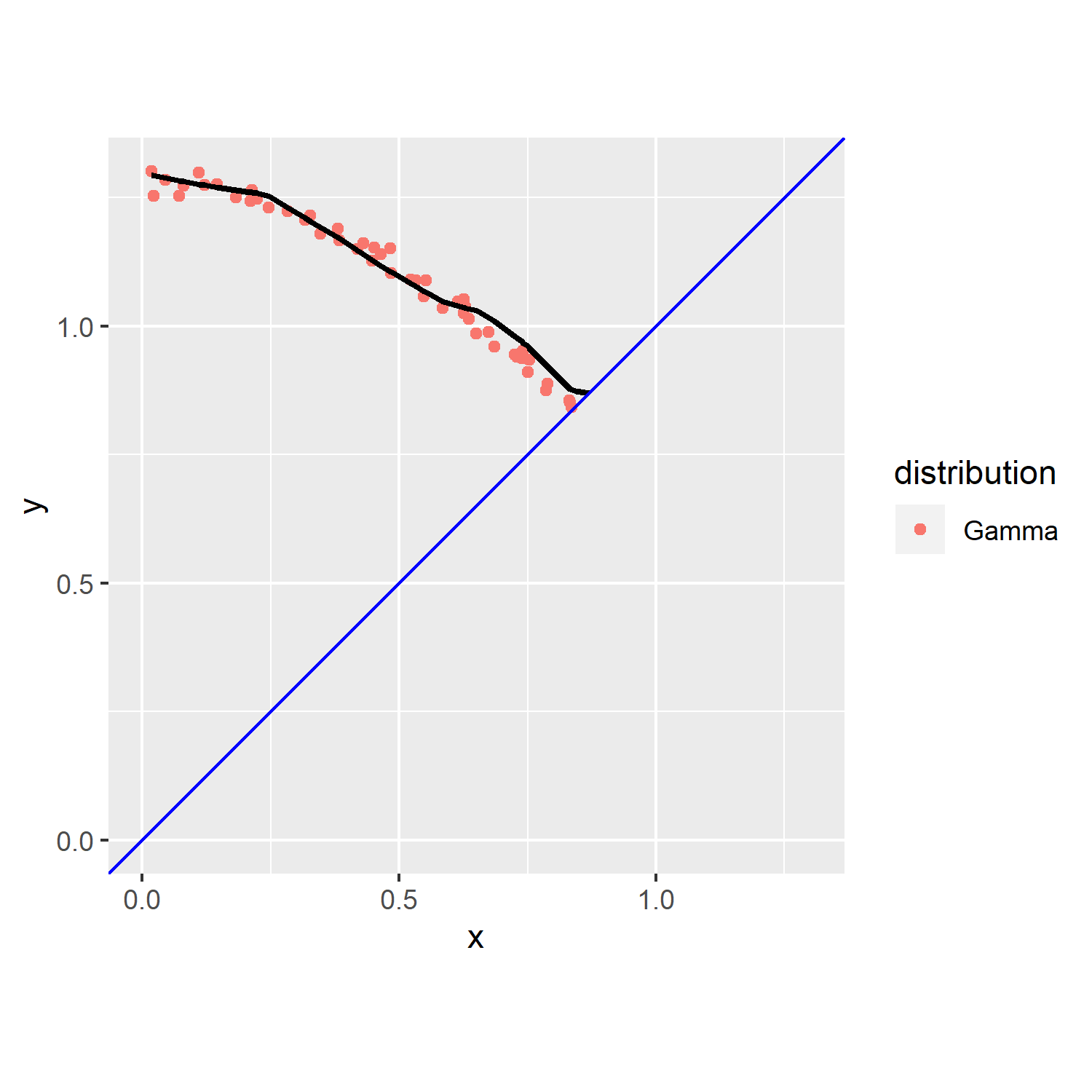}}\\
	\subfloat[Beta Distribution $\mu_e = 33.68$]{\includegraphics[width=.375\textwidth]{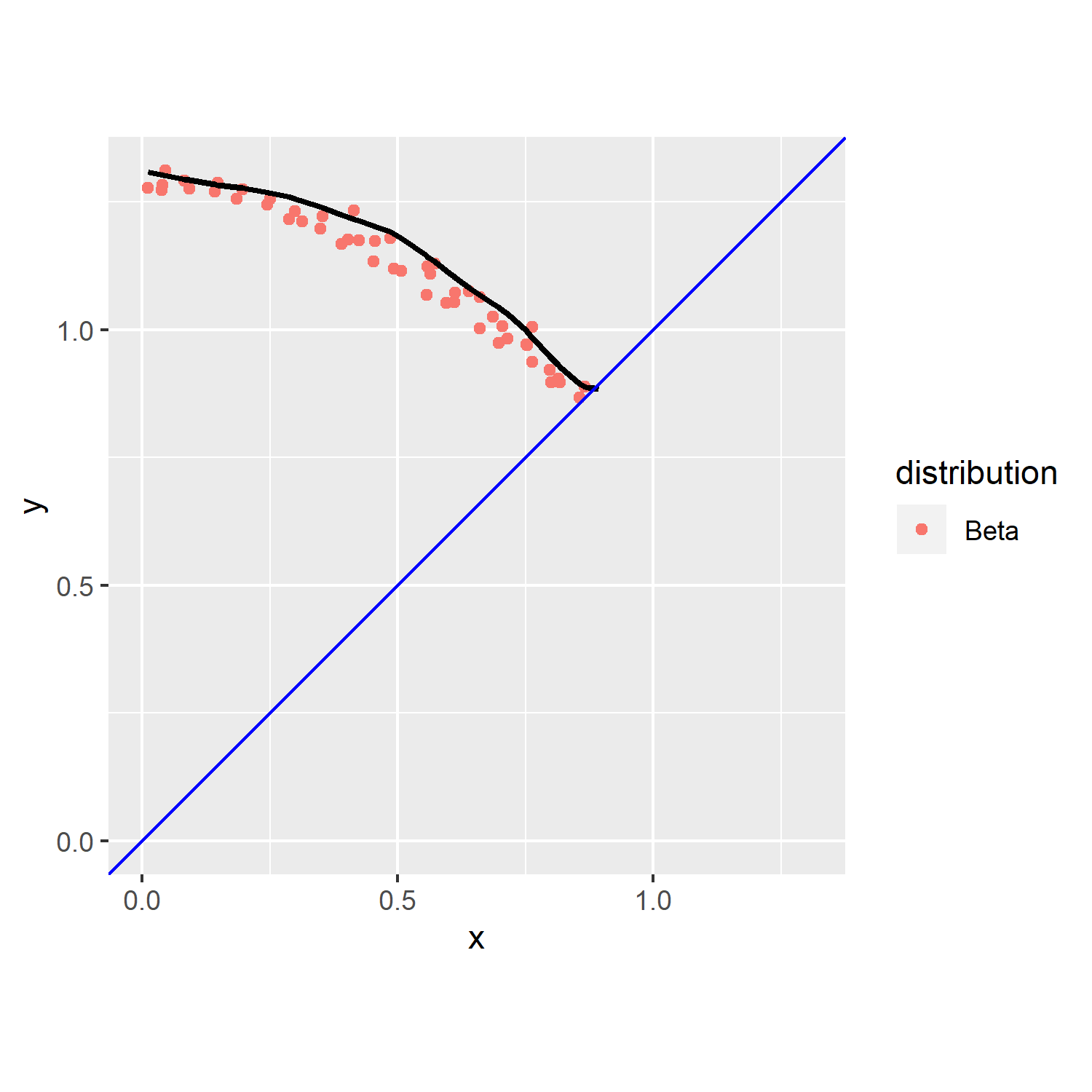}}\quad
	\caption{Illustration of neural network model predicting shape for \textit{test data}, for three representative distributions}
	\label{test}
\end{figure}

It is not straightforward to visualize the model performance, but it is a good idea to plot model predictions against simulated values for common passage time distributions, and to do this for both training and test data. In Figure \ref{train} and \ref{test} we plot model predictions against simulated values for a \textit{representative} passage time distribution within each distribution family (normal, gamma, or beta). The concept of a representative passage time distribution is connected to how the simulations are performed, and we refer to Section \ref{simulations} for details --- still with a representative distribution time we mean: the passage time distribution whose mean equals the median of all simulated passage time means within that distribution family. 

The rest of the note is structured as follows. In Section \ref{modelinterlude} we introduce neural networks and the underlying theory. In Section \ref{simulations} we go through how the simulations are performed and which transformations are made to data.
Finally, in Section \ref{results} we explain how the models are evaluated, and in Section \ref{conclusion} we summarize the results.

\section{Model Interlude}
\label{modelinterlude}
In this section we give a short overview of the theory behind neural networks.

\subsection{Neural Networks and Deep Learning}
\label{nn}
Simply put, neural networks is a class of functions capable of approximating a bigger class of functions --- namely all Lebesgue integrable functions. The models are governed by a set of parameters called the \textit{weights} ($W$) and \textit{biases} ($b$), where the weights have the same role as the coefficients in a linear regression, and the bias that of the intercept.

There are three types of neural networks, all with different structure and \textit{primary} use: \textit{feed-forward} networks for regression, \textit{convolutional} networks for image classification, and \textit{recurrent} networks for natural language processing. For our purposes we shall only need feed-forward neural networks, and henceforth refer to them as simply neural networks. First there is a distinction between a neural \textit{network} and a neural network \textit{model}. The former is a type of function, while the latter is a statistical model (that is, a parameterized family of probability distributions) capable of approximating a large class of functions. We begin by defining a feed-forward network.

A neural network with parameters $\theta = \{(W^{(i)}, b^{(i)}),\ i=1,2,\ldots,l+1 \}$ is a (typically non-linear) function $f_{\theta}: \mathbb{R}^k \to \mathbb{R}^m$ satisfying
\begin{enumerate}[$(i)$]
	\item $f_{\theta}$ is a composition: $f_{\theta}(x) = f^{(l+1)}(f^{(l)}(\ldots (f^{(1)}(x )))) $.
	\item $f_{\theta}$ is a \textit{special} form of composition: $h^{(0)} = x$, $h^{(i)}:= f^{(i)}(h^{(i-1)}) = g^{(i)}(W^{(i)}h^{(i-1)}+b^{(i)}  ) $, where $g^{(i)}$ is an \textit{activation} function (see $(iii)$) applied component-wise to the vector $W^{(i)}h^{(i-1)}+b^{(i)}$ (i.e. $W^{(i)}$ is a matrix, and $h^{(i)}$, $b^{(i)}$ are vectors).
	\item $g^{(i)}$ is an \textit{activation} function, which means than that $g^{(i)}$ belongs to a list of functions which the machine learning community currently defines as activation functions. At the moment, the most popular activation functions are
\begin{table}[H]
	\begin{tabular}{c|c}
		Name & $g(z)$ \\ \hline
		ReLu & $\max(0,z)$ \\
		Elu & $\max(\alpha (e^z-1) ,z)$ \\
		Leaky ReLu & $\max(\alpha,z)$ \\
		tanh & $\tanh(z)$ \\
		Sigmoid & $1/(1+e^{-z})$
	\end{tabular}
\end{table}	\noindent
Usually, the same activation functions is used for all layers, except for the output layer, i.e. $g^{(i)}(z) = g(z)$ for $i=1,\ldots,l$.
\end{enumerate}
We use the following nomenclature for the parameters of the network:
\begin{itemize}
	\item $W^{(i)} = [W^{(i)}_{k,j}]_{k,j}, i=1,\ldots, l$, are called the \textit{weights}.
	\item $b^{(i)} = (b^{(i)}_1,\ldots, b^{(i)}_{n_i})^t, i=1,\ldots, l$, are called the \textit{biases}.
	\item $h^{(i)} = (h^{(i)}_1,\ldots, h^{(i)}_{n_i} )^t, i=1,\ldots, l$, are called the \textit{hidden layers}.
	\item $n_i$, the length  of the vectors $b^{(i)}$ and $h^{(i)}$, is called the \textit{width} of the $i$th hidden layer.
	\item $l$ is the number of hidden layers.
\end{itemize}
Note that $n_i$, $i = 1,\ldots, l$ determine the dimensions of the matrices $W^{(i)}$ and the vectors $b^{(i)}$.
An example will help to make the concepts clearer. 
\subsubsection{Example --- Forward-pass}
Let $\tilde{y} =  f_{\theta}(x_1,x_2,x_3,x_4,x_5)$ be a single-layer neural network function with parameters $\theta = \{(W^{(1)},b^{(1)}),(W^{(2)}b^{(2)}) \}$, as illustrated in Figure 2.
Assume that $f_{\theta}$ has ReLu activation, i.e $g(z) = \max\{0,z\}$ and that the output is continuous, e.g. a regression model. In order to go from input to  output we do the following calculations
\begin{enumerate}
	\item $h^1 = g(W^1x+b^1)\  (\text{component-wise}) \implies h_i^1 = \max\{0, \sum_{j=1}^{4}W^1_{ij}x_j+b^1_i   \}$
	\item $\tilde{y} = W^2h^1+b^2 = W^2_1h^1_1+W^2_2h^1_2+W^2_3h^1_3+b^2$.
\end{enumerate}
\input{forwardpass.tex}

\subsubsection{Universal Approximation Theorem}
Neural networks have found great success in a variety of situations, see e.g. \cite{Lin_2017}, and it is worth spending a few paragraphs reflecting on why this is the case. The following result is a good starting point.

\begin{theorem}{\cite[Thm. 1]{UAT}}
	For any Lebesgue integrable function $g:\ \mathbb{R}^n\to \mathbb{R}$ and $\epsilon > 0$ there exists a fully-connected ReLu feed-forward neural network $f_{\theta}$ of width $d_m \leq n+4$ such that
	\begin{align*}
		\int_{\mathbb{R}^n}|g(x)-f_{\theta}(x)|dx < \epsilon.
	\end{align*}
\end{theorem}

For practical purposes, the class of Lebesgue integrable functions contains basically all functions we could possibly be interested in, and therefore the result is quite astonishing at first glance. It tells us that, for a given function and desired accuracy, there exists a neural network able to approximate it.
However, the result does not give much away modeling-wise. For instance, it tells us very little about how wide the network needs to be, and nothing about how deep (how many hidden are layers needed). 

The result goes someway in explaining the success of neural nets, but there are many classes of functions with similar approximating capabilities which have \textit{not} found great modeling success. For instance, hige degree polynomials (through the Stone–Weierstrass theorem), or just the class of all continuous functions. Clearly, just approximating ability is not enough. We want the class to be able to approximate a large class of functions to be interesting, but we also need:
\begin{enumerate}[$(i)$]
	\item The approximating class of functions should be small enough to be "searchable", i.e. it must be feasible to find a good approximating function from the class.
	\item There has to exists a search algorithm on the class of functions, i.e. we must be able to search the class in an efficient and statistically sound way.
\end{enumerate}
In addition to the universal approximation theorem, neural networks typically satisfy $(i)$ and $(ii)$. The algorithm commonly used to train neural networks (solving $(ii)$) is called \textit{backpropagation} (in combination with \textit{stochastic gradient descent}). Still, there is no clear answer as to why neural networks has worked so well in applications, but the above arguments goes someway in explaining it on a general level. Another explanation for their success, albeit less technical, can perhaps be found in the name, \textit{neural} network. Neural networks are \textit{inspired} by the structure of the human brain, and the argument goes: it is therefore plausible that they should be good at solving problems the human brain excels at solving. 

\subsubsection{Neural networks for regression}
Let $(x_1,y_1),\ldots, (x_n,y_n)$ be a set of observations with $x_i \in \mathbb{R}^k$ and $y_i \in \mathbb{R}$. Write $x,y$ for the whole set of data. We assume the following model for the data: $$y_i = f_{\theta}(x_i)+\epsilon_i$$ where $f_{\theta}$ is a neural network acting as the mean of the distribution, and $\{\epsilon_i \}$ is the randomness, often taken to be i.i.d. N$(0, \sigma^2)$ (but not in this note). In our case $f_{\theta}(x)$ would represent the true shape and $\epsilon$ the random deviation that occurred due to simulation. Note that in this context $x$ is a point in the input space (i.e. not just the value of the $x$-coordinate) and is therefore a vector composed of the mean, standard deviation, and percentiles of the passage time as well as the value of the $x$-coordinate. 

The likelihood is denoted by $\mathbb{P}_{\theta}(y|x)$ (in this context $\mathbb{P}_{\theta}(y|x)$ is just notation, and has nothing to do with a Bayesian approach) The likelihood is made explicit with assumptions on the distribution of $\{\epsilon_i\}$. 
The statistical model is given by
\begin{align*}
	\mathcal{P} = \{  \mathbb{P}_{\theta}(y|x);\ \theta \in \Theta  \}
\end{align*}
where $\Theta$ is the parameter space. We call $\mathcal{P}$ a neural network regression model, i.e. a class of probability distributions, where the mean of each observation is determined by a neural network.

Statistical inference on this model entails picking an \textit{optimal} parameter $\theta = \{(W^i,b^i)\} \in \Theta$. This is achieved by minimizing a \textit{loss function}, and common loss functions include:

\begin{table}[H]
	\begin{tabular}{c|c}
		Name & Loss $L(x,y;\theta)$ \\ \hline
		Mean Square Error & $\frac{1}{n}\sum_{i=1}^{n} (y_i-f_{\theta}(x_ i))^2 $ \\ 
		Mean Absolute Error & $\frac{1}{n}\sum_{i=1}^{n} |y_i-f_{\theta}(x_ i)| $ \\
		Mean Absolute Percentage Error & $\frac{100}{n}\sum_{i=1}^{n} \frac{|y_i-f_{\theta}(x_ i)|}{y_i} $
	\end{tabular}
\end{table}
Note that for these loss functions it is not necessary to have an explicitly defined likelihood. This simplifies matters, but at the cost of not being able to make any distributional based inference (e.g. p-values, confidence intervals et cetera). There are of course other loss functions making this possible (e.g. the negative log likelihood), but here we are only concerned with predicting $\mathcal{B}$, and not making any further inference.

\subsubsection{Pros and Cons of a Feed-forward Network}
What are pros and cons with using a neural network as basis for a statistical model in comparison to traditional models, e.g. linear regression? Some are listed below.\\
\textbf{Pros:}
\begin{itemize}
	\item If $y$ has non-linear dependencies on $x$, this can be modeled \textit{indirectly} by the network (if data is plentiful), whereas in traditional models this has to be modeled \textit{directly}, e.g. by adding an $x^2$-term.
	\item The same holds true for interactions --- they are modeled indirectly by the network, whereas in e.g. linear regression this has to be modeled explicitly.
\end{itemize}
\textbf{Cons:}
\begin{itemize}
	\item Not enough data will lead to overfitting, and poor performance.
	\item Traditional models, and linear regression in particular, are much more \textit{interpretable} in that they can be used to get insight into the problem at hand. Usually, this is not the case for neural networks, who's primary use is prediction, not interpretation.
\end{itemize}
For our problem, we see that a neural network is a suitable model choice: we can simulate data, so data is plentiful; we are mainly interested in predicting the shape, not explaining it; and we have very little a priori knowledge of the functional relationship between the shape and passage time distribution.
\section{Simulations}
\label{simulations}

Our dataset consists of 240.000 simulations with passage times from three different families of random variables: normal, gamma, and (scaled) beta. 

Each simulation begins by uniformly choosing which of the three families the passage times should belong to. The parameters of the passage time distribution are chosen \textit{once} in the beginning of each simulation. Then the parameters are chosen as follows:
\begin{itemize}
	\item Normal distribution $\tau_e\sim$ N$(\mu, \sigma^2)|\tau_e > 0$: simulate $\mu\sim U(1,19)$, $\sigma \sim U(1,10)$.
	\item Gamma distribution $\tau_e\sim\Gamma(n, \lambda)$: simulate $\lambda \sim \frac{1}{U(1,3)}$, $n\sim \text{ceiling}(U(1,9))$, repeat until $ \frac{n}{\lambda} \leq 20$.
	\item Beta distribution $\tau_e\sim a\cdot B(\alpha, \beta)$: simulate $a \sim U(2,40)$, $\alpha \sim U(0.5, 5)$, $\beta \sim  U(0.5, 5)$, repeat until $\frac{a \alpha}{\alpha + \beta} \leq 20.$
\end{itemize}
This means that the mean $\mu_e$ of a simulated passage time will fall in the interval $[1 , 20]$, and that $\mu_e \approx 10$. All simulations run until 300.000 edges have been traversed.

Later, when we test the predictive power of the model, we will predict shapes on distributions which have expected value in the interval $[25,45]$, i.e. we will check if the model generalizes well to completely new data. The test data is generated as follows:
\begin{itemize}
	\item Normal distribution $\tau_e\sim$ N$(\mu, \sigma^2)|\tau_e > 0$: simulate $\mu\sim U(25,45)$, $\sigma \sim U(10, 20)$.
	\item Gamma distribution $\tau_e\sim\Gamma(n, \lambda)$: simulate $\lambda \sim \frac{1}{U(5,9)}$, $n\sim \text{ceiling}(U(1,9))$,
	repeat until $ 25 \leq \frac{n}{\lambda} \leq 45$.
	\item Beta distribution $\tau_e\sim a\cdot B(\alpha, \beta)$: simulate $a \sim U(25,100)$, $\alpha \sim U(0.5, 5)$, $\beta \sim  U(0.5, 5)$, repeat until $25 \leq \frac{a \alpha}{\alpha + \beta} \leq 45$
\end{itemize}

Regarding the shapes generated by the beta distribution we note that the parameter $a$ simply acts as a scaling on the shape, i.e. if $\tau_e \sim B(\alpha, \beta)$ generates the shape $\mathcal{B}$ then $\tau_e \sim a \cdot B(\alpha, \beta)$ generates the shape $a \cdot \mathcal{B}$. We have still included this parameter, since we wanted a broad distribution family with bounded support where both the shape and the location can be varied. Furthermore, since we condition on $\frac{a \alpha}{\alpha + \beta} \leq 20$ in the training data and $25 \leq \frac{a \alpha}{\alpha + \beta} \leq 45$ in the test data, the parameters $a, \alpha, \beta$ are not independent so the test data is \textit{not} simply a rescaling of the training data.

\subsection{Transformation of Data}
\label{transformation}
\begin{figure}[H]
	\centering
	\subfloat[][\ \ \ \ \ \ Full Simulation]{\includegraphics[width=.4\textwidth]{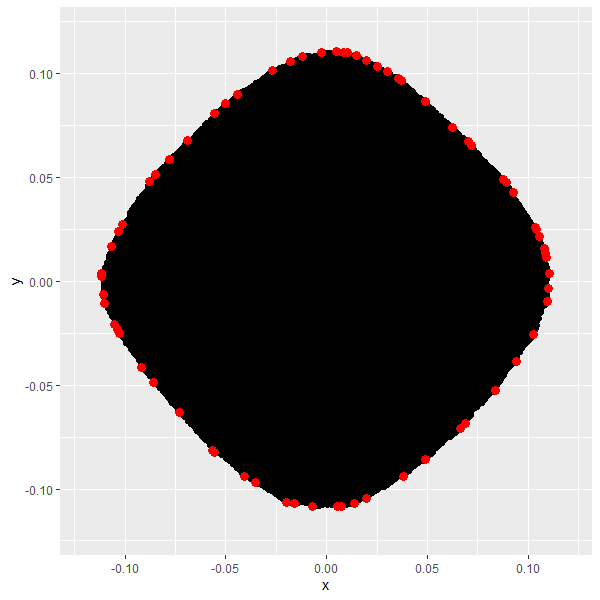}}\quad
	\subfloat[][\ \ \ \ \ \ Convex Hull]{\includegraphics[width=.4\textwidth]{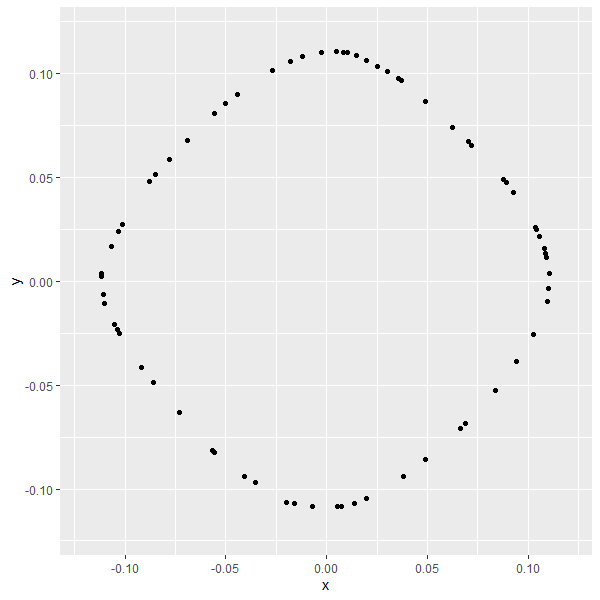}}\\
	\subfloat[][\ \ \ \ \ \ First Quadrant Projection]{\includegraphics[width=.4\textwidth]{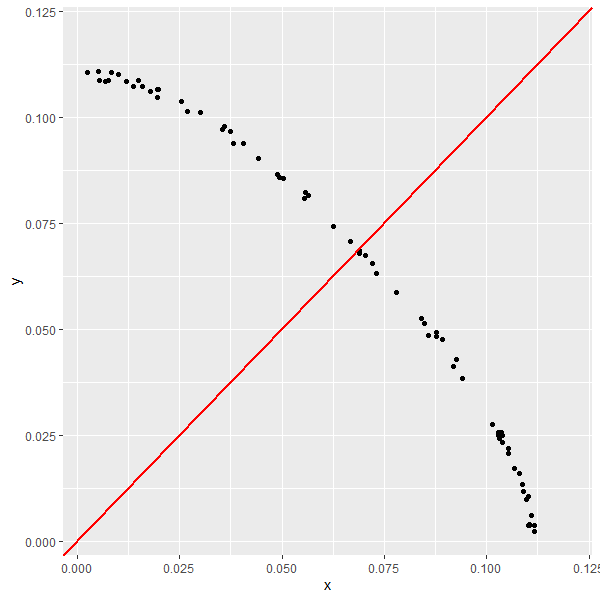}}\quad
	\subfloat[][\ \ \ \ \ \ Reflection in the line $y = x$]{\includegraphics[width=.4\textwidth]{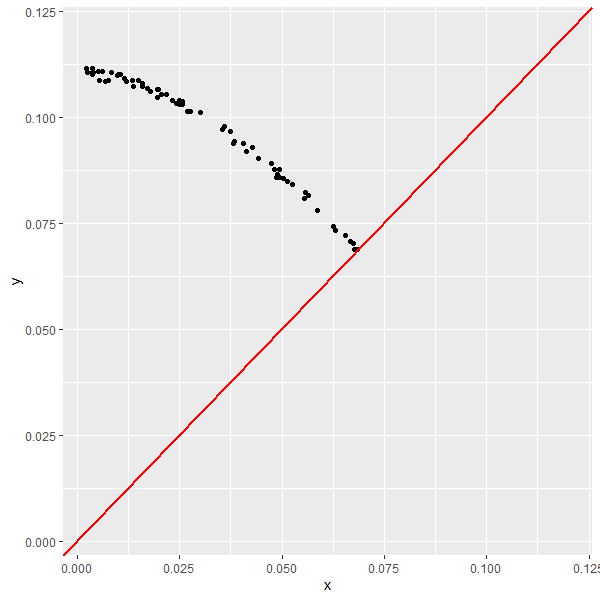}}
	\caption{Illustration of transformations used on data.}
	\label{fig2}
\end{figure}

As mentioned, not much is known about the asymptotic shape $\mathcal{B}$. But we do know that the shape will be 1.) convex and compact, and 2.) inherit all the symmetries of $\mathbb{Z}^2$. We of course want our models to reflect this as much as possible. Convexity and compactness can not be ensured beforehand with a neural network, and instead has to be learned from data. Since 2.) holds it is enough if the model can predict $\mathcal{B}$ the part of the first quadrant that lies above the line $y = x$.
With this in mind we make the following transformations to our simulated data:

\begin{enumerate}
	\item The complete data is replaced with its convex hull $\{(x,y) \}$ --- the smallest convex set containing $B(t)$ (also compact).
	\item The convex hull is projected onto the first quadrant through $\{(x,y) \} \to \{(|x|,|y|)\}$.
	\item The points in this set that lie below the line $y = x$ are reflected in the same line.
\end{enumerate}
Figure \ref{fig2} illustrates the effect of the transformations on a simulation.
\noindent The resulting dataset will be referred to as \textit{raw data}.

\begin{figure}[H]
	\centering
	\subfloat[(A) Boxplot for all $y$-values for data with transformation (1)-(3)]{\includegraphics[width=.4\textwidth]{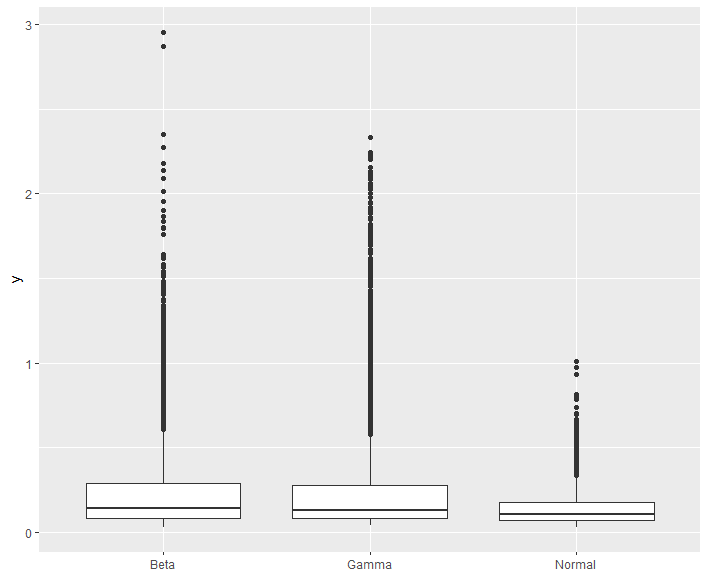}}\quad
	\subfloat[(B) Boxplot for all $y$-values for mean-transformed data]{\includegraphics[width=.4\textwidth]{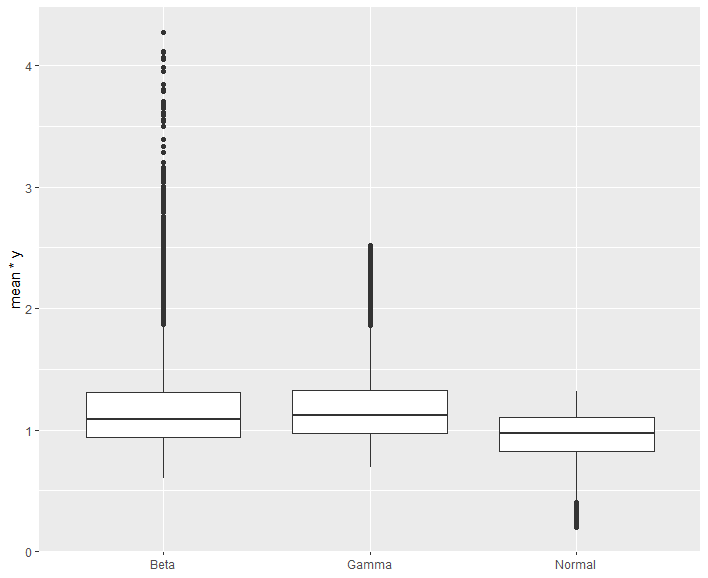}}\\
	\subfloat[(C) $\mu_e$ plotted against $\max(y)$ for raw data]{\includegraphics[width=.4\textwidth]{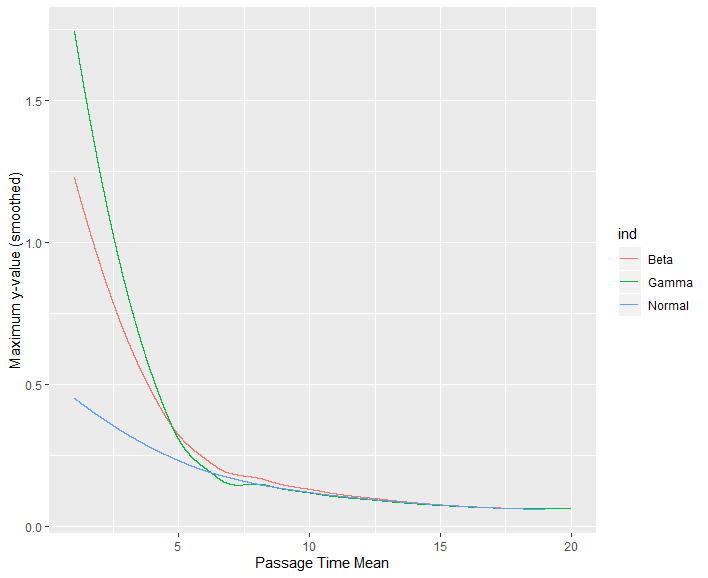}}\quad
	\subfloat[(D) $\mu_e$ plotted against $\max(\mu_e \cdot y)$ (mean-transformed data)]{\includegraphics[width=.4\textwidth]{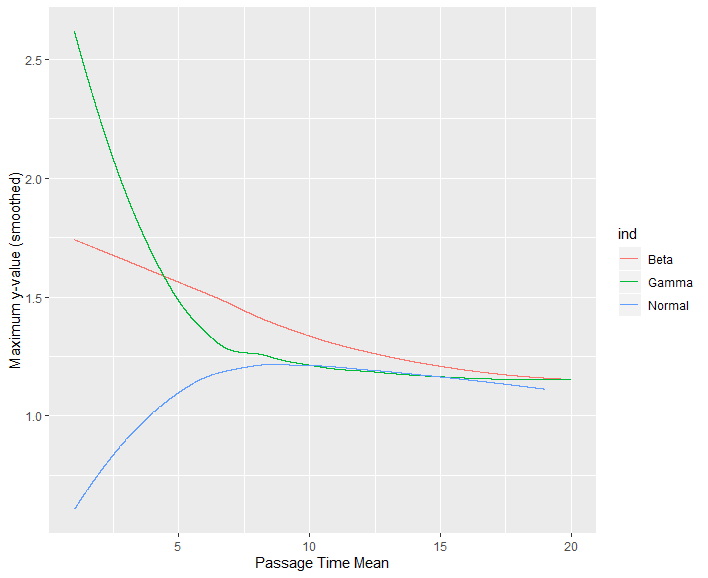}}
	\caption{}
	\label{fig4}
\end{figure}

Furthermore, in Figure \ref{fig4} (A) and (C) we can see that the data is rather skewed, with some outliers present. The skewness in data comes from the fact that passage times with larger means \textit{tend} to generate a "smaller" shape. This is illustrated in Figure \ref{fig4} (C) where the passage time mean is plotted against the largest $y$-value for each distribution family, where for visualization purposes a smoothing function is applied to the $y$-values.

For modeling purposes it might be a good idea to mitigate this. A new dataset called \textit{mean-transformed data} is constructed by multiplying all $(x,y)$-values with their corresponding mean of the passage time, i.e.
$$(x,y) \to (\mu_e x, \mu_e y).$$

Since smaller shapes tends to correspond to larger passage time means, this transformation has the effect that smaller shapes tends to become larger, and larger shapes tends to become smaller. Note, that this transformation has to do only with modeling purposes, not any a priori knowledge. 

The resulting dataset --- mean-transformed data --- is more uniform in the shape sizes, which can be seen in Figure \ref{fig4} (B) and (D). Our hope is that this data might be more suitable for modeling. However, we will fit models to both datasets, to investigate which one that forms the best basis for a prediction model. 

\section{Results}
\label{results}

\subsection{Evaluation Approach}
Each model candidate is fitted to the datasets \textit{raw} and \textit{mean-transformed}, and evaluated according to the metrics listed below. In order to select the final model, each model candidate is evaluated using cross-validation. Cross-validation is a common tool used for model selection, and for assessing how a given statistical model will generalize to \textit{new} data.
In this note, cross-validation is used as a model selecting tool, and we asses how models will generalize by trying them on a test dataset. 
The cross-validation procedure is as follows.
\begin{enumerate}
	\item The training data is randomly partitioned into $k$ folds. 
	\item For each fold, the fold is removed from the data and the model fitted to the remaining data. This model is then used to predict values for the excluded fold, and some error metric is calculated.
	\item Summarize the error metrics for the folds, e.g. by taking the mean of the metrics.
\end{enumerate}
The models considered here are evaluated using a 10-fold cross-validation with the errors calculated according to the metrics in the table below.
\begin{itemize}
	\item Mean Absolute Error (\textit{mae}) :  $\frac{\sum_{i=1}^{n} |y_i-\hat{y}_i|}{n}$
	\item Mean Absolute Percentage Error (\textit{mape}):  $ \sum_{i=1}^{n} \frac{|y_i-\hat{y}_i|}{y_i} / n $
\end{itemize}
As final model, we selected the one with the lowest cross-validation error. 

Furthermore, the models are also tested on a completely new dataset, where we use the mean absolute percentage error as a metric --- recall that for both training and cross-validation data we have $\mu_e \in [1,20]$.
The test data consists of 120.000 simulations where $\mu_e \in [25,45]$, and the data have been generated following the schema outlined in Section \ref{simulations}. Hence, this evaluation approach also tests if the model can generalize to new data.
Each model is summarized with the above listed metrics on the training data, cross-validation data, and the prediction data.

Note that we do not use common overfitting reducing methods such as Lasso, ridge regression, and dropout. Overfitting seems not to be a problem here --- since data is too plentiful. Overfitting tend to be a problem when model complexity is greater than data complexity, and the latter we can increase with more data. We did however also test the most common methods for reducing overfitting (Lasso and ridge regression, as well as dropout) and found that they \textit{reduced} model performance.

\subsection{Base Line Models --- Linear Regression}
Using linear regression we fit the models listed in Table \ref{table:reg1} and \ref{table:reg2}. They are fitted to data following standard methods for minimizing a Gaussian likelihood (minimizing the mean squared error). The models are meant to serve as a simple base comparison for the neural network models. 

\subsubsection{A word on Notation}
Recall that the input to the models consists of the following variables: $\mu,\sigma,x, q_{0.01}, \ldots, q_{0.99}$, i.e. the mean, the standard deviation, an $x$-coordinate, and the percentiles.
We use the following short-hand notation in denoting the models:
\begin{itemize}
	\item $\textbf{q} = \{ q_{0.01}, \ldots, q_{0.99} \}$.
	\item \textbf{all} = $\{ \mu,\ \sigma,\ x, q_{0.01}, \ldots, q_{0.99} \}$ = all variables.
	\item (colon) \textbf{:} denotes standard \textit{pair-wise} interaction term, e.g. $x : q_{0.01}$ denotes the interaction between $x$ and $q_{0.01}$ and $x : \textbf{all} $ denotes the pair-wise interaction between $x$ and all other variable.
\end{itemize}
	
	\begin{table}[H]
		\centering
		\begin{tabular}{clcccc}
			\hline
			data\_source & model\_formulas & mae\_train & mape\_train & mape\_cv & mape\_test \\ 
			\hline
			raw & $y \sim \textbf{all}+x:\textbf{all}+x^2:\textbf{all}$ & 0.052 & 35.53 & 35.44 & 748.10 \\ 
			raw & $y\sim \mu_e+\sigma_e+x+x^2$ & 0.09 & 55.52& 55.3273419 & 1437.25 \\ 
			raw & $y\sim \textbf{all}+x^2 $ & 0.091 & 57.53 & 57.28 & 1438.98 \\ 
			\hline
		\end{tabular}
	\caption{Performance metrics the for regression models on \textit{raw data} dataset}
	\label{table:reg1}
	\end{table}
	
	\begin{table}[H]
		\centering
		\begin{tabular}{clcccc}
			\hline
			data\_source & model\_formulas & mae\_train & mape\_train & mape\_cv & mape\_test \\ 
			\hline
			mean\_transformed & $y \sim \textbf{all}+x:\textbf{all}+x^2:\textbf{all}$ & 0.15 & 12.90 & 12.88 & 43.28 \\ 
			mean\_transformed & $y\sim \textbf{all}+x^2 $ & 0.14 & 13.29 & 13.22 & 40.64 \\ 
			mean\_transformed & $y\sim \mu_e+\sigma_e+x+x^2$ & 0.22 & 21.93 & 21.82 & 41.89 \\ 
			\hline
		\end{tabular}
		\caption{Performance metrics the for regression models on \textit{mean-transformed data} dataset}
	\label{table:reg2}
	\end{table}
\noindent
Clearly, the mean-transformed data is better suited as basis for modeling.

\begin{figure}[H]
	\centering
	\subfloat[Normal Distribution $\mu_e = 10.03$]{\includegraphics[width=.35\textwidth]{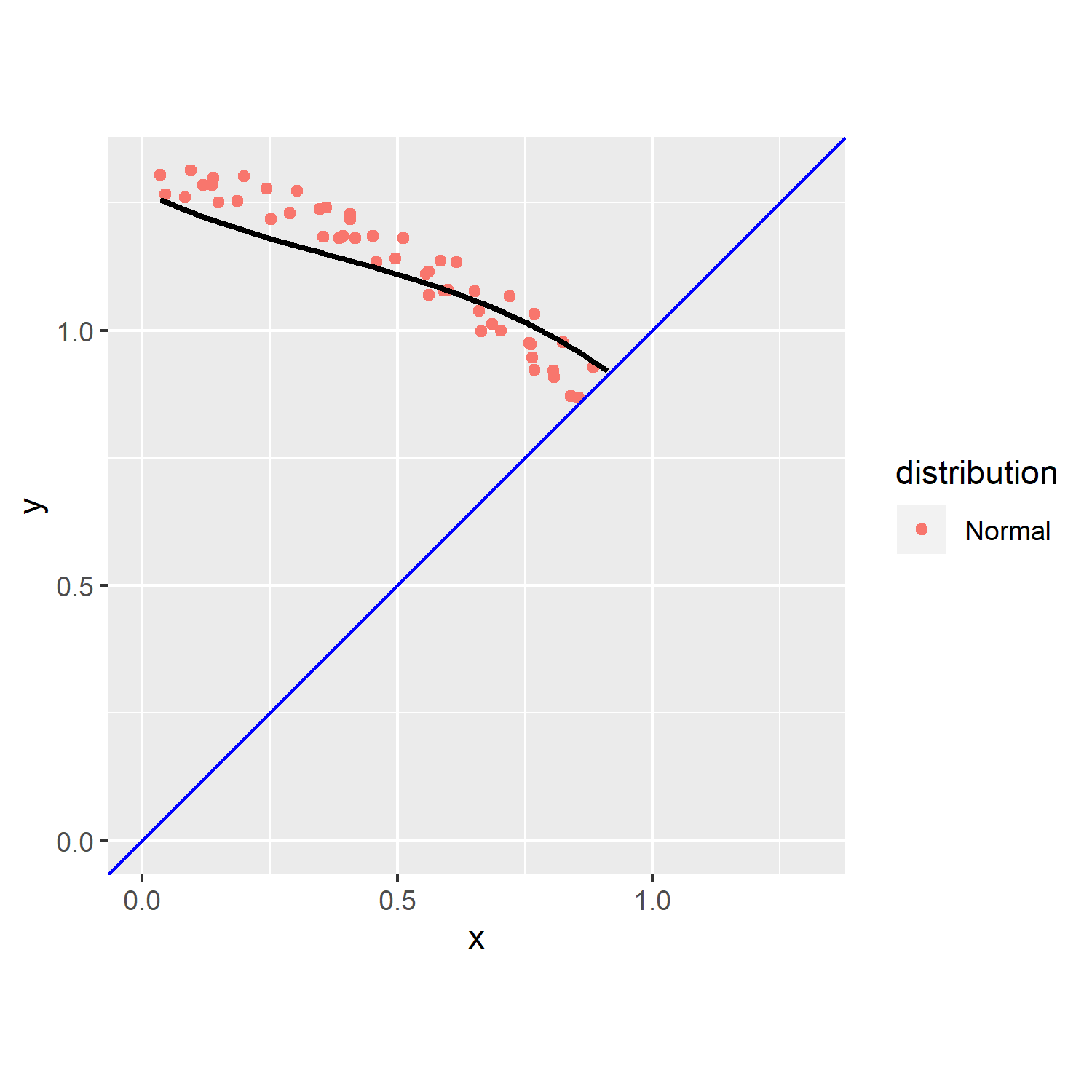}}\quad
	\subfloat[Gamma Distribution $\mu_e = 8.32$]{\includegraphics[width=.35\textwidth]{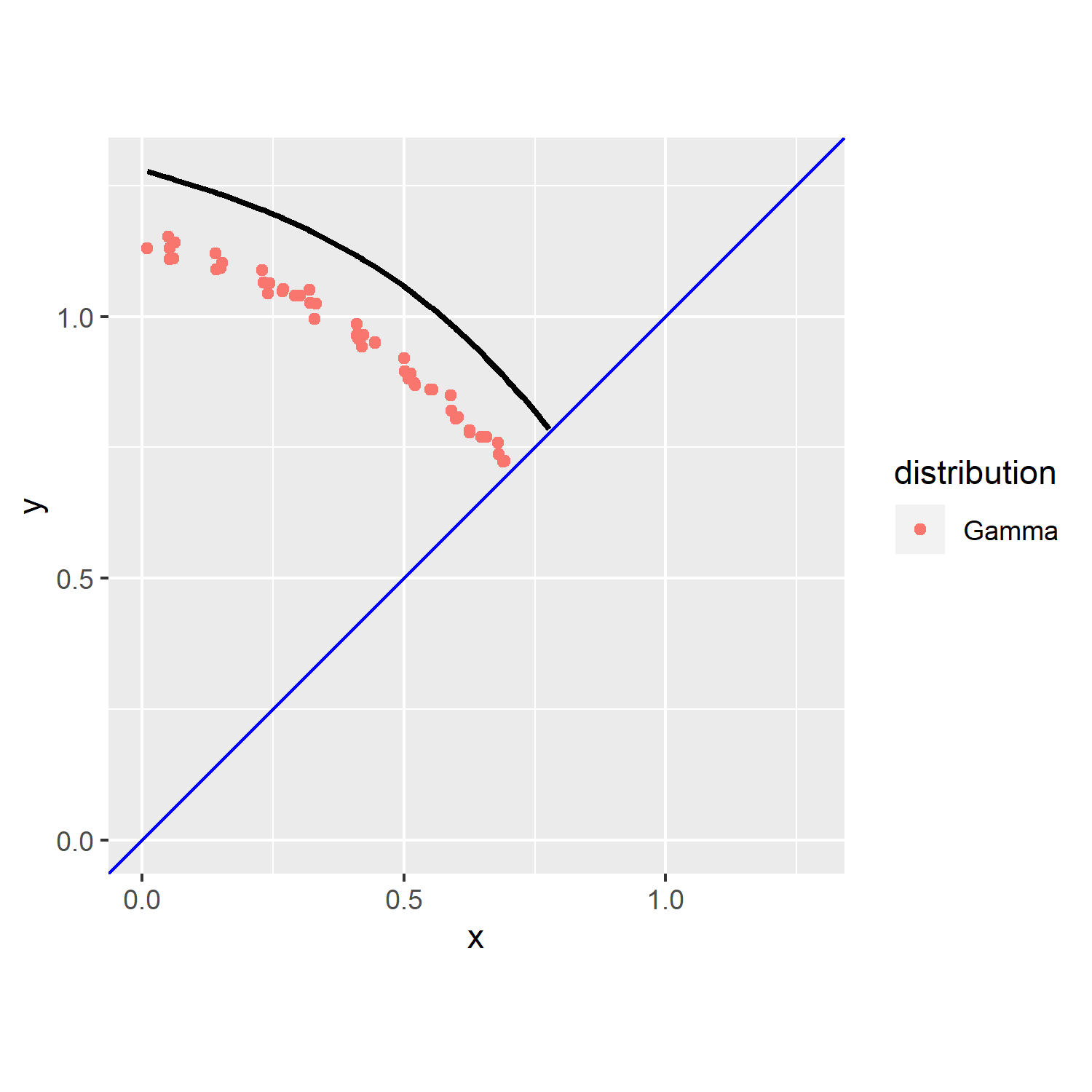}}\\
	\subfloat[Beta Distribution $\mu_e = 8.39$]{\includegraphics[width=.35\textwidth]{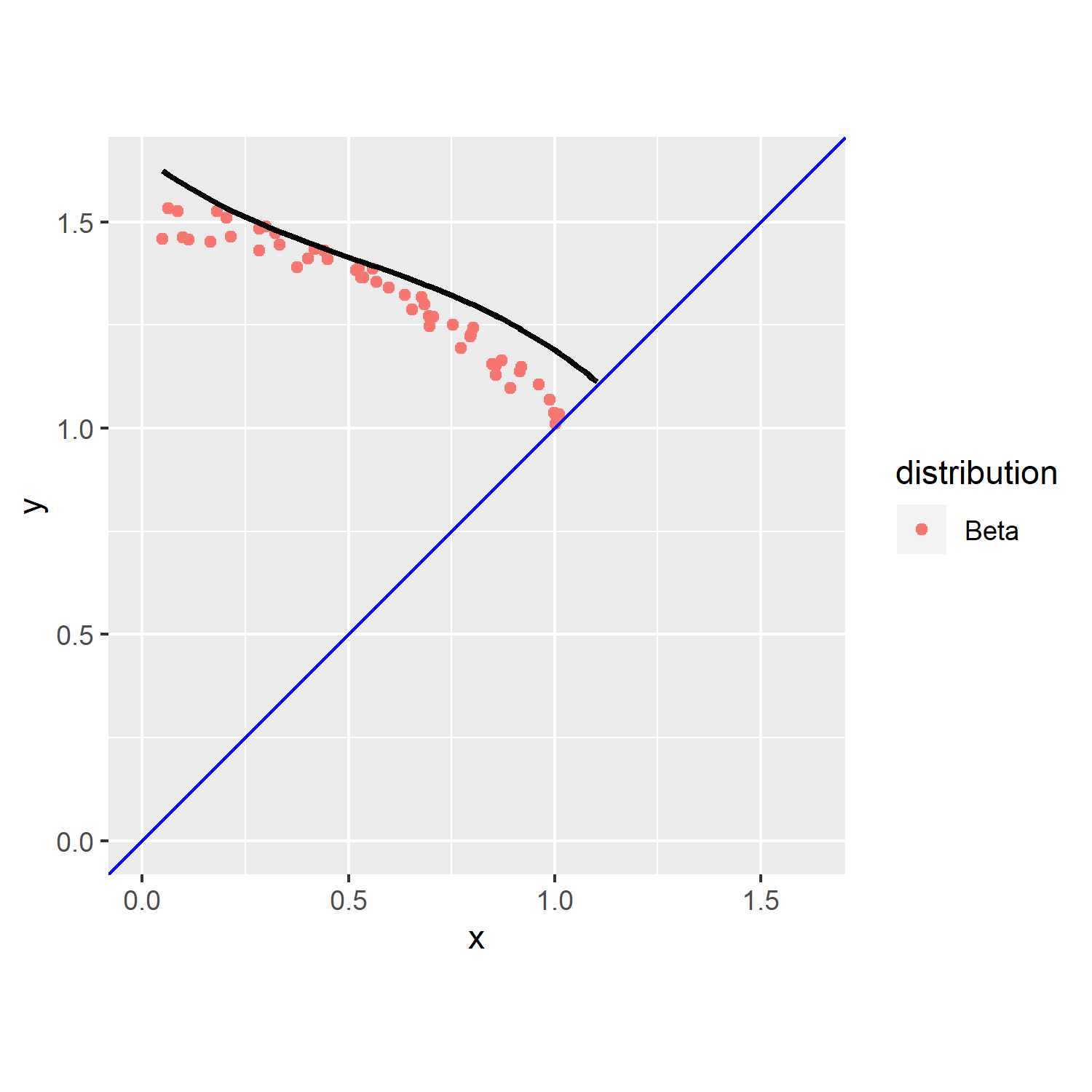}}\quad
	\caption{Illustration of regression model predicting shape for \textit{mean-transformed training} data, for three distributions with mean equal to median mean of data}
	\label{fig6}
\end{figure}

\begin{figure}[H]
	\centering
	\subfloat[Normal Distribution $\mu_e = 35.04$]{\includegraphics[width=.35\textwidth]{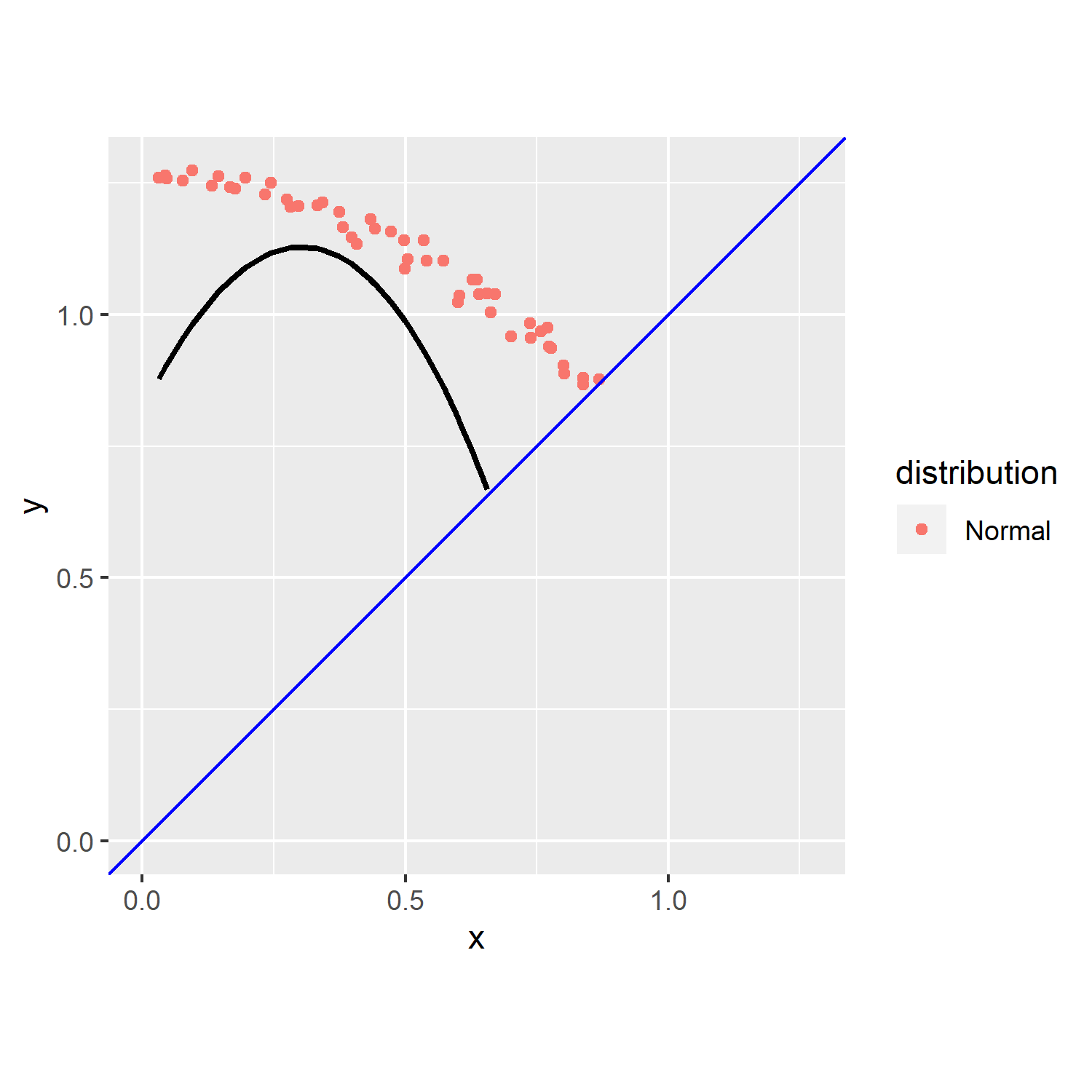}}\quad
	\subfloat[Gamma Distribution $\mu_e = 34.97$]{\includegraphics[width=.35\textwidth]{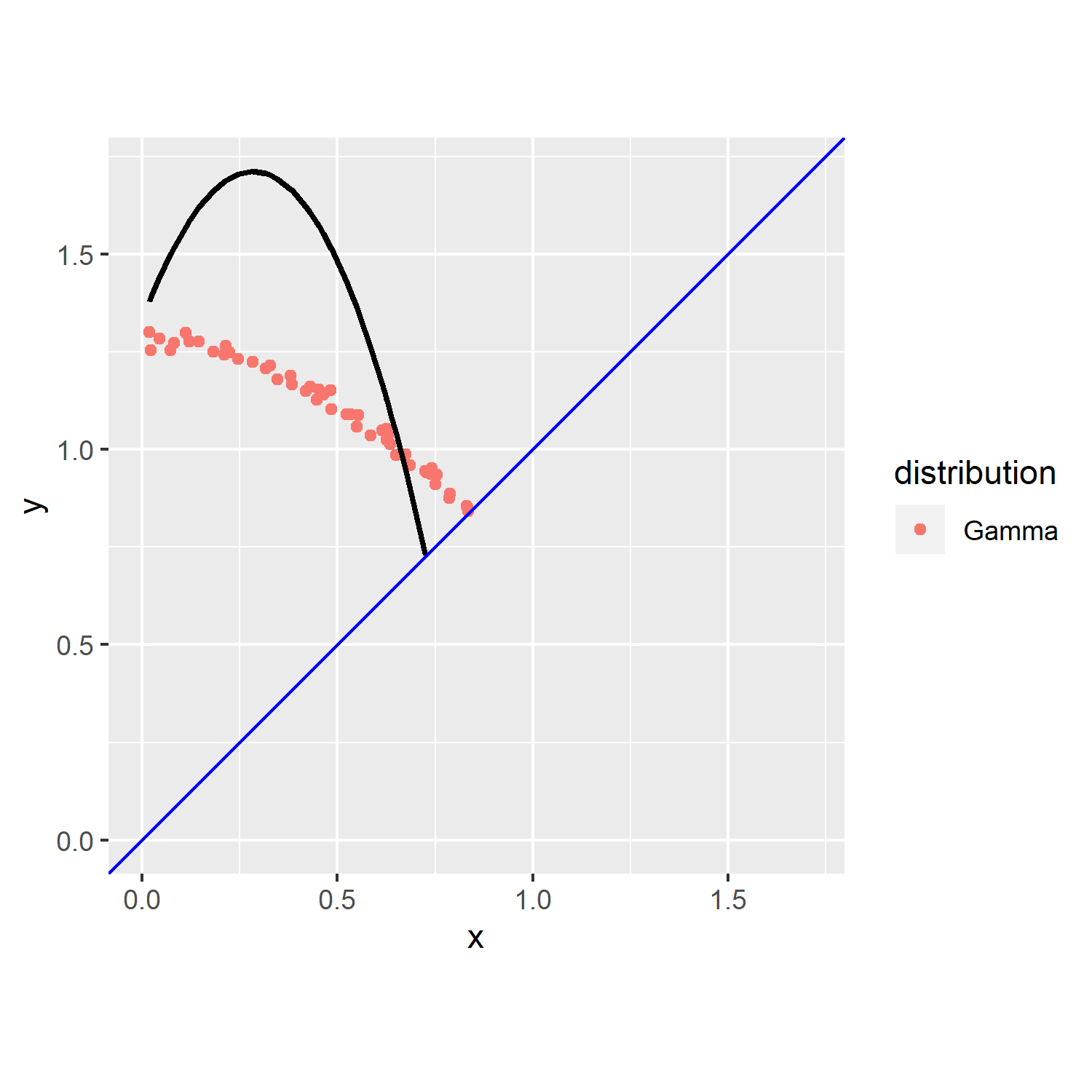}}\\
	\subfloat[Beta Distribution $\mu_e = 33.68$]{\includegraphics[width=.35\textwidth]{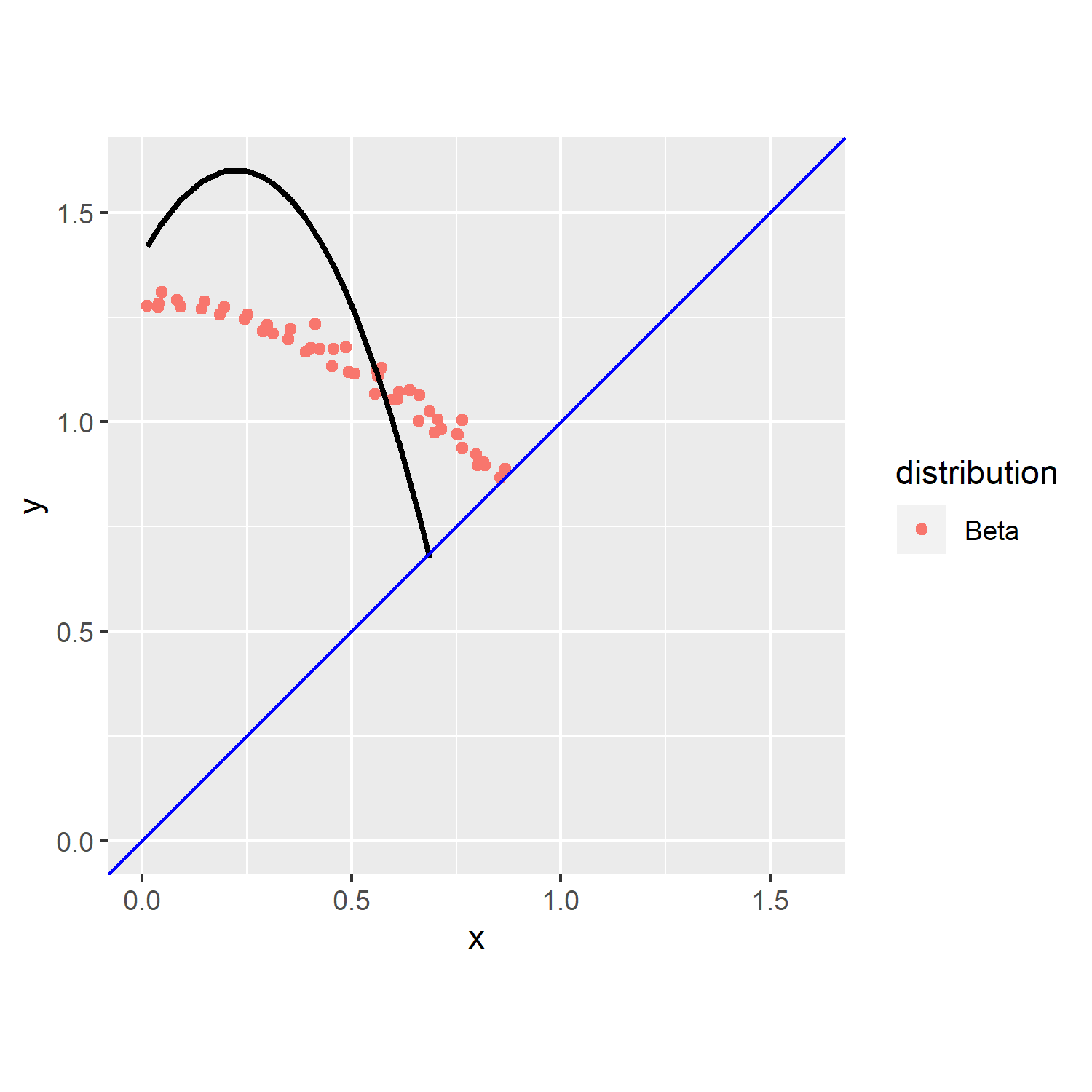}}\quad
	\caption{Illustration of regression model predicting shape for \textit{mean-transformed test} data, for three distributions with mean equal to median mean of data}
	\label{fig7}
\end{figure}

Figure \ref{fig6} and \ref{fig7} illustrates how the final linear regression model predicts on training and test data. For each distribution we have picked the simulation where the passage time mean equals the median of all passage time means for the distribution class --- i.e. the mean for which $\mu_e = \text{median}(\mu_{i,d},\ i=1,\ldots, 240.000/3)$ where $\mu_{i,d}$ denotes the passage time mean of the $i$th simulation for the distribution $d$ (gamma, normal, or beta).

\textit{These} regression models does not predict new shapes well. It is clear that the non-linearity introduced in the models by adding an $x^2$-term is the wrong kind of non-linearity. It works well on training data, but does not generalize well. It should of course be possible to improve on these models following standard modeling practices. Again, the regression models are only meant to serve as a base comparison for the neural network models, and very little statistical craftsmanship has been applied to the models.

\newpage
\subsection{Neural Network Models}

To each dataset we fit 12 different neural networks: all combinations of $4,6,8,10$ layers and $40,60,80$ hidden units.
All networks uses ReLu activation, and are trained by minimizing the mean absolute percentage error.

The next paragraph contains technicalities and details about how the networks have been trained and may be skipped.
The networks are trained using stochastic gradient descent with adaptive moment estimation (\textit{adam}), and the mean absolute error as the loss function. A batch size of 64 is used an each model gets two epochs, with the final model re-trained using five epochs. An R code snippet for training a model:
\begin{verbatim}
model %>% compile(
  loss = "mean_absolute_percentage_error", 
  optimizer = optimizer_adam(lr = 0.001, beta_1 = 0.9, beta_2 = 0.999,
  epsilon = NULL, decay = 0, amsgrad = FALSE, clipnorm = NULL,
  clipvalue = NULL)
  )
	
history <-  model %>% 
fit(x = train_data, y = train_label, 
    batch_size = 64, 
    epochs = 2, 
    validation_split = 0.1)
\end{verbatim}

The results are presented in Tables 3 and 4, and in Figures \ref{train} and \ref{test} . As with regression we see that the mean-transformed data is more suitable to base the models on. All models have better predictive performance on this dataset, according to the mean absolute percentage error. 

\begin{table}[H]
	\centering
	\begin{tabular}{lrrrrrr}
		\hline
		data\_source & model\_layers & nr\_hidden\_units & mae\_train & mape\_train & mape\_cv & mape\_pred \\ 
		\hline
		raw & 8.00 & 60.00 & 0.01 & 3.13 & 1.92 & 35.94 \\ 
		raw & 10.00 & 60.00 & 0.01 & 3.21 & 2.25 & 30.34 \\ 
		raw & 4.00 & 80.00 & 0.01 & 3.07 & 2.27 & 32.96 \\ 
		raw & 8.00 & 40.00 & 0.01 & 3.19 & 2.60 & 29.86 \\ 
		raw & 6.00 & 60.00 & 0.01 & 3.23 & 2.97 & 14.71 \\ 
		raw & 4.00 & 40.00 & 0.01 & 3.23 & 3.00 & 28.10 \\ 
		raw & 6.00 & 40.00 & 0.01 & 3.22 & 3.07 & 64.77 \\ 
		raw & 10.00 & 40.00 & 0.01 & 3.16 & 3.13 & 17.61 \\ 
		raw & 10.00 & 80.00 & 0.01 & 3.24 & 3.28 & 22.77 \\ 
		raw & 8.00 & 80.00 & 0.01 & 3.14 & 3.54 & 14.87 \\ 
		raw & 4.00 & 60.00 & 0.01 & 3.17 & 4.01 & 24.99 \\ 
		raw & 6.00 & 80.00 & 0.01 & 3.10 & 5.55 & 26.43 \\ 
		\hline
	\end{tabular}
	\caption{Performance metrics the for neural networks models on \textit{raw data} dataset}
\label{table:nn1}
\end{table}

\begin{table}[H]
	\centering
	\begin{tabular}{lrrrrrr}
		\hline
		data\_source & model\_layers & nr\_hidden\_units & mae\_train & mape\_train & mape\_cv & mape\_pred \\ 
		\hline
		mean\_transformed & 10.00 & 60.00 & 0.02 & 1.72 & 1.52 & 3.09 \\ 
		mean\_transformed & 6.00 & 80.00 & 0.02 & 1.71 & 1.53 & 2.91 \\ 
		mean\_transformed & 8.00 & 80.00 & 0.02 & 1.72 & 1.56 & 2.17 \\ 
		mean\_transformed & 10.00 & 40.00 & 0.02 & 1.74 & 1.56 & 2.77 \\ 
		mean\_transformed & 8.00 & 60.00 & 0.02 & 1.72 & 1.65 & 3.52 \\ 
		mean\_transformed & 8.00 & 40.00 & 0.02 & 1.74 & 1.68 & 3.25 \\ 
		mean\_transformed & 10.00 & 80.00 & 0.02 & 1.72 & 1.70 & 3.07 \\ 
		mean\_transformed & 6.00 & 40.00 & 0.02 & 1.75 & 1.70 & 2.83 \\ 
		mean\_transformed & 4.00 & 40.00 & 0.02 & 1.79 & 1.70 & 3.56 \\ 
		mean\_transformed & 6.00 & 60.00 & 0.02 & 1.71 & 1.73 & 3.69 \\ 
		mean\_transformed & 4.00 & 60.00 & 0.02 & 1.74 & 1.74 & 4.30 \\ 
		mean\_transformed & 4.00 & 80.00 & 0.02 & 1.73 & 1.98 & 3.44 \\ 
		\hline
	\end{tabular}
		\caption{Performance metrics the for neural networks models on \textit{mean-transformed} dataset}
\label{table:nn2}
\end{table}

As the final model we pick the network with lowest cross validation error, which would be the 10 layer, 60 hidden units network.

\subsubsection{Performance for Pareto Distributions}
As a final test of the model's generalizing ability we test its predictive performance on shapes generated by Pareto distributed edge times. This test data consists of 12.000 simulations where the parameters of the Pareto distribution are chosen as follows:
\begin{itemize}
	\item (the shape) $\alpha \sim U(2,7)$
	\item (the scale/location) $x_m \sim U(1,46)$
\end{itemize}
This dataset has also been mean-transformed in the same fashion as before. This results in a mean absolute percentage error of $4.9\%$ on this dataset. 
\begin{figure}[H]
	\includegraphics[scale=0.6]{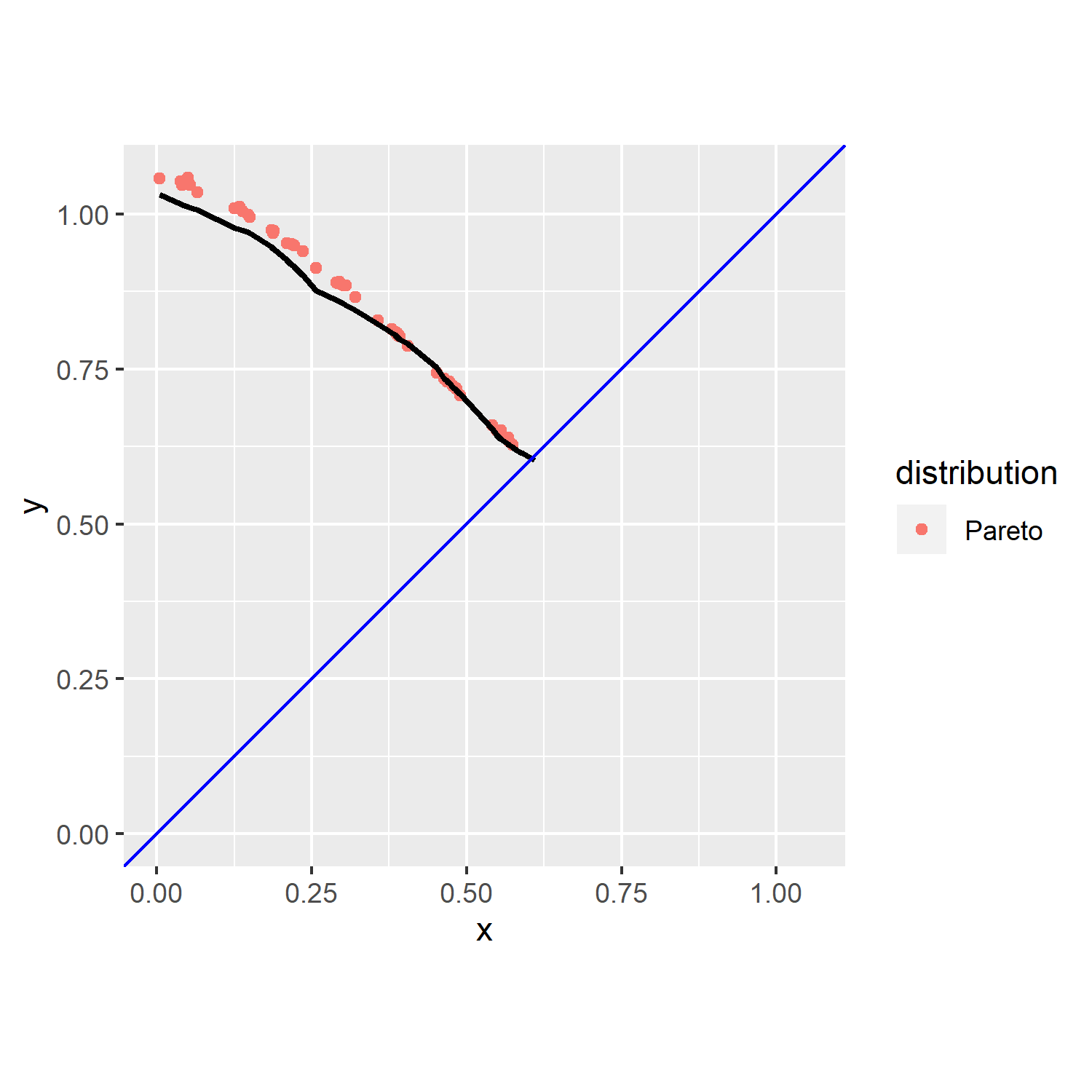}
	\caption{Pareto Distribution $\mu_e = 35.1$}
\end{figure}

\section{Conclusion}
\label{conclusion}
We have seen that is it possible to use a neural network for predicting the shape $\mathcal{B}$ for passage times belonging to a normal, gamma, or beta distribution. Furthermore, the model seems to generalize fairly well to new data --- very well for data with passage times belonging to the same distribution families as the simulated data (but with parameter values from a different regime). The generalization to new data with passage times belonging to a Pareto distribution is slightly less good, but the error produced (4.9\%) is still lower than the regression model errors on training and test data. 

The results gives us hope that it should be possible to construct better neural nets, able to predict $\mathcal{B}$ for a larger number of distribution families. However, it is not our expectation that there will be a \textit{single} neural network able to predict $\mathcal{B}$ for any passage time distribution.

We expect the models to improve by simulating more data, from new passage time distributions. In this note we used neural nets with a single value as output, i.e. given $x,\mu, \sigma, q_{0.01}, \ldots, q_{0.99} $ the model produces a single $y$-value. This is not the only approach, one could construct neural nets with vector valued outputs. For instance, a net that maps $\mu, \sigma, q_{0.01}, \ldots, q_{0.99}$ to $\{(x_i, y_i)\}$. This would allow for different loss functions, e.g. the total distance.

\section{Acknowledgments}
I would like to thank my supervisor Professor Mia Deijfen for introducing me to the model, and for helpful comments and thoughts on the manuscript.

\bibliography{articleRef}{}
\bibliographystyle{siam}

\end{document}

%% file: forwardpass.tex
\begin{figure}[H]
	\label{fig:nn}
	
	\centering
	
	\begin{tikzpicture}[
	clear/.style={
		draw=none,
		fill=none,
	},
	net/.style={
		matrix of nodes,
		nodes={
			draw,
			circle,
			inner sep=10pt
		},
		nodes in empty cells,
		column sep=2cm,
		row sep=-19pt
	} 
	]
	\matrix[net] (mat)
	{
		|[clear]| \parbox{1.3cm}{\centering Input\\layer} 
		& |[clear]| \parbox{1.3cm}{\centering Hidden\\layer} 
		& |[clear]| \parbox{1.3cm}{\centering Output\\layer} \\
		
		$x_{1}$  & |[clear]|        & |[clear]| \\
		|[clear]|         & $h_{1}^{1}$ & |[clear]| \\
		$x_{2}$  & |[clear]|        & |[clear]| \\
		|[clear]|         & |[clear]|        & |[clear]| \phantom{$\alpha_{0}^{0}$} \\
		$x_{3}$  & $h_{2}^{1}$ & $\tilde{y}$ \\
		|[clear]|         & |[clear]|        & |[clear]|  \phantom{$\alpha_{0}^{0}$} \\
		$x_{4}$  & |[clear]|        & |[clear]| \\
		|[clear]|         & $h_{3}^{1}$ & |[clear]| \\
		$x_{5}$  & |[clear]|        & |[clear]| \\ 
	};

	\foreach \ai in {2,4,...,10}
	\draw[<-] (mat-\ai-1) -- +(-2cm,0);
	
	\foreach \ai in {2,4,...,10} {
		\foreach \aii in {3,6,9}
		\draw[->] (mat-\ai-1) -- (mat-\aii-2);
	}
	
	\foreach \ai in {3,6,9}
	\draw[->] (mat-\ai-2) -- (mat-6-3);
	
	\draw[->] (mat-6-3) -- node[above] {Output} +(3cm,0);
	
	\end{tikzpicture}
	
	\caption{Illustration of $f_{\theta}(x_1,\ldots, x_5)$}
	
\end{figure}